\pdfoutput=1

\documentclass[11pt]{article}

\usepackage[]{acl}

\usepackage{times}
\usepackage{latexsym}

\usepackage[T1]{fontenc}

\usepackage[utf8]{inputenc}

\usepackage{microtype}

\usepackage{amsmath}
\usepackage{amsfonts}
\usepackage{graphicx}
\usepackage{float}
\usepackage{subfigure}
\usepackage{bm}
\usepackage{multirow}
\usepackage{color}

\title{TreeMAN: Tree-enhanced Multimodal Attention Network for ICD Coding}

\author{Zichen Liu$^{\dagger}$ \quad Xuyuan Liu $^{\dagger}$ \quad Yanlong Wen$^{\dagger }$ \thanks{\ \ Corresponding author.}  \quad Guoqing Zhao $^{\ddagger}$\quad  Hongbin Wang$^{\ddagger}$\quad Xiaojie Yuan$^{\dagger}$  \\
$^{\dagger}$TKLNDST, College of Computer Science, Nankai University
\\$^{\ddagger}$Mashang Consumer Finance Co.,Ltd.
\\
\{liuzichen,wenyl,yuanxj\}@dbis.nankai.edu.cn\\
hsuyuanliu0204@mail.nankai.edu.cn\\
\{guoqing.zhao02,hongbin.wang02\}@msxf.com
}

\begin{document}
\maketitle

\begin{abstract}
  ICD coding is designed to assign the disease codes to electronic health records (EHRs) upon discharge, which is crucial for billing and clinical statistics. 
  In an attempt to improve the effectiveness and efficiency of manual coding, many methods have been proposed to automatically predict ICD codes from clinical notes.
  However, most previous works ignore the decisive information contained in structured medical data in EHRs, which is hard to be captured from the noisy clinical notes.
  In this paper, we propose a \textbf{Tree}-enhanced \textbf{M}ultimodal \textbf{A}ttention \textbf{N}etwork (TreeMAN) to fuse tabular features and textual features into multimodal representations by enhancing the text representations with \textit{tree-based features} via the attention mechanism.
  \textit{Tree-based features} are constructed according to decision trees learned from structured multimodal medical data, which capture the decisive information about ICD coding.
  We can apply the same multi-label classifier from previous text models to the multimodal representations to predict ICD codes.
  Experiments on two MIMIC datasets show that our method outperforms prior state-of-the-art ICD coding approaches. 
\end{abstract}

\section{Introduction}

The International Classification of Diseases (ICD), maintained by the World Health Organization, is a hierarchical classification of codes representing diseases, injuries, and so on.
ICD codes have been used in diverse areas, including insurance reimbursement, epidemiology, and clinical research \citep{park2000accuracy}.

In the hospital, when patients discharge, their electronic health records (EHRs) and all associated data are transferred to the information management department, where clinical coders manually assign the appropriate ICD codes using rigid ICD coding guidelines after reviewing records \citep{o2005measuring}.
The manual code assignment is expensive, labor-intensive, and error-prone due to the large volume of medical record information and high professional requirements \citep{DBLP:conf/amia/NguyenTKKCOZKHL18}.

Since deep learning has achieved great success in lots of healthcare applications \citep{DBLP:journals/access/CaiWLL19},  many neural methods have been proposed to automate the ICD coding process by researchers \citep{teng2022review}.
Recent works formulate automated ICD coding as a multi-label document classification task, using clinical notes as model input, predicting coding with a multi-label classifier, and learning text features through word embedding techniques and neural networks such as RNNs and CNNs \citep{DBLP:conf/naacl/MullenbachWDSE18,DBLP:conf/ijcai/VuNN20,DBLP:conf/acl/ZhouC000NCL20}.
To improve the code representation learning, researchers further leverage features of ICD codes such as hierarchical structures \citep{DBLP:conf/acl/CaoCLZLC20} and descriptions \citep{DBLP:conf/naacl/MullenbachWDSE18,DBLP:conf/acl/ZhouC000NCL20}.
However, most previous methods ignore structured medical data, including physiological data collected by medical sensors and medical record information such as prescriptions and microbiology test results in EHRs.
The few methods that leverage structured data are either ensemble-based approaches \citep{DBLP:conf/mlhc/XuLPGBMPKCMXX19} or data-mining methods that discard semantic information \citep{ferrao2021can}. 

\begin{figure}[t]
  \centering
  \includegraphics[width=\columnwidth]{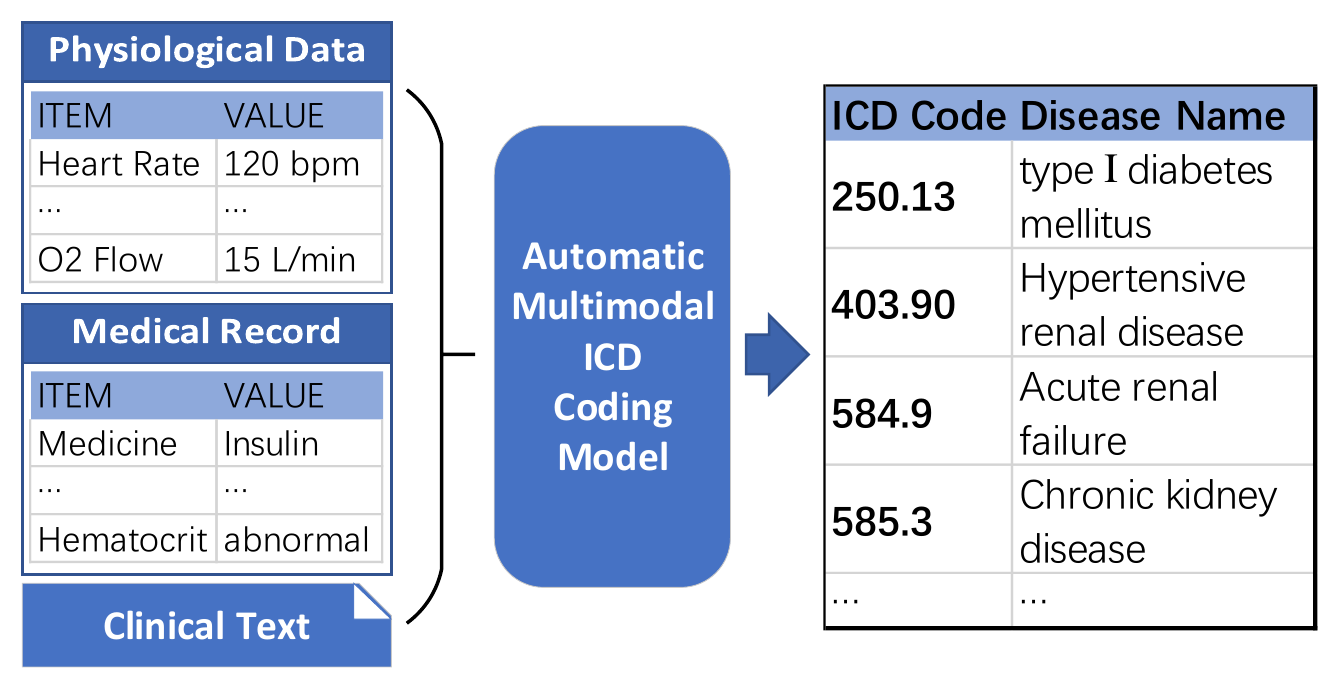}
  \caption{An example of automatic multimodal ICD coding. Model inputs include physiological data and medical records in addition to clinical text.}\label{fig:intro}
\end{figure}

In this work, we argue that structured medical data can improve coding accuracy by enhancing semantic representations and providing more information because clinical notes are noisy and ambiguous.
For example, there are many different types of insulin, like ``Insulin Aspart'' and ``Insulin Glargine'', which are often written the same in notes but clearly distinguished by Generic Sequence Numbers (GSNs) in medical records. 
Considering different writing styles and polysemous abbreviations, predicting ICD codes from clinical notes is more complicated.
However, automatic multimodal ICD coding (as Figure \ref{fig:intro} shown) is challenging for the following reasons: 
1) medical data is naturally heterogeneous, with data types including numerical quantities, categorical values, and derived time series such as perioperative vital sign signals \citep{DBLP:conf/acl/ZhouC000NCL20}; 
2) the feature selection method needs to be designed especially for multi-ICD codes as it's a multi-label classification task; 
3) decisive information for a code in the long clinical note may be contained in short segments that are likely different for different codes \citep{DBLP:conf/naacl/MullenbachWDSE18}.



In this paper, 
we propose a novel \textbf{Tree}-enhanced \textbf{M}ultimodal \textbf{A}ttention \textbf{N}etwork named TreeMAN 
to address the aforementioned problems. 
Since it's hard to do feature engineering for structured medical data, 
we construct \textit{tree-based features} from the structured medical data through decision trees that require little data preparation \citep{DBLP:journals/tsmc/SafavianL91} 
instead of manually crafting features based on medical knowledge.
Inspired by previous works \citep{DBLP:conf/www/Wang0FNC18,DBLP:conf/kdd/HePJXLXSAHBC14},
we represent the tree-based features by embedding vectors.
Taking the tree-based embeddings and text representations as input, TreeMAN applies an attention mechanism to select relevant tree-based features for text representations and output fused multimodal representations that contain richer information to benefit the downstream classifier. 
However, our method has limitations in handling long-tailed labels as it is difficult to build a decision tree from less than 10 positive samples. 

\paragraph{Contributions.} 
In summary, the main contributions of our work include:
\begin{itemize}
  \item We propose a multimodal ICD coding framework that exploits structured medical data in EHRs to construct tree-based features to enhance text representations.
  \item We propose TreeMAN, a tree-enhanced multimodal attention network, which fuses text representations and tree-based features into unified multimodal representations by the attention mechanism. To the best of our knowledge, it's the first model to jointly learn multimodal features for the ICD coding task.
  \item Experiments demonstrate the effectiveness of our proposed method. Results on two datasets show that TreeMAN outperforms previous state-of-the-art ICD coding methods.
\end{itemize}

\paragraph{Reproducibility:} Source code will be released upon paper acceptance.

\section{Related Work}\label{sec:related}

\subsection{ICD Coding}

Research on Automatic ICD coding can be traced back to nearly 30 years ago when \citet{DBLP:conf/sigir/LarkeyC96} proposed an ensemble algorithm to integrate different types of classifiers to assign ICD codes to inpatient discharge summaries. 
A series of methods based on Deep Neural Networks has been implemented on this task since this paradigm achieved colossal success in Clinical NLP.
\citet{perotte2014diagnosis} built ``hierarchical'' Support  Vector  Machines (SVMs) outperforming the "flat" classifier. 
\citet{DBLP:conf/naacl/MullenbachWDSE18} built a convolutional attention model which combined the single filter CNN module and the per-label attention module. 
A series of network modules based on attention mechanism have been utilized after the early attempts, including multi-scale attention module \citep{DBLP:conf/cikm/XieXYZ19},  residual convolution module \citep{DBLP:conf/aaai/Li020}. 
We also notice that the hierarchical structure of ICD-9 could be effectively described by a joint-classification module on different levels \citep{DBLP:conf/ijcai/VuNN20} or in the form of specific hyperbolic representation \citep{DBLP:conf/acl/CaoCLZLC20}.

Multimodal learning methods help to integrate multiple information like test reports, nursing notes, etc., in the MIMIC-III datasets. 
An early attempt was made by \citep{DBLP:conf/mlhc/XuLPGBMPKCMXX19}, in which an ensemble-based approach was developed to integrate the structured and unstructured text of different modalities. 
\citet{DBLP:conf/pkdd/RajendranZSP21} made full use of unstructured information by effectively exploiting the geometric properties of pre-trained word embeddings.

\begin{figure*}[h]
  \centering
  \includegraphics[width=0.91\textwidth]{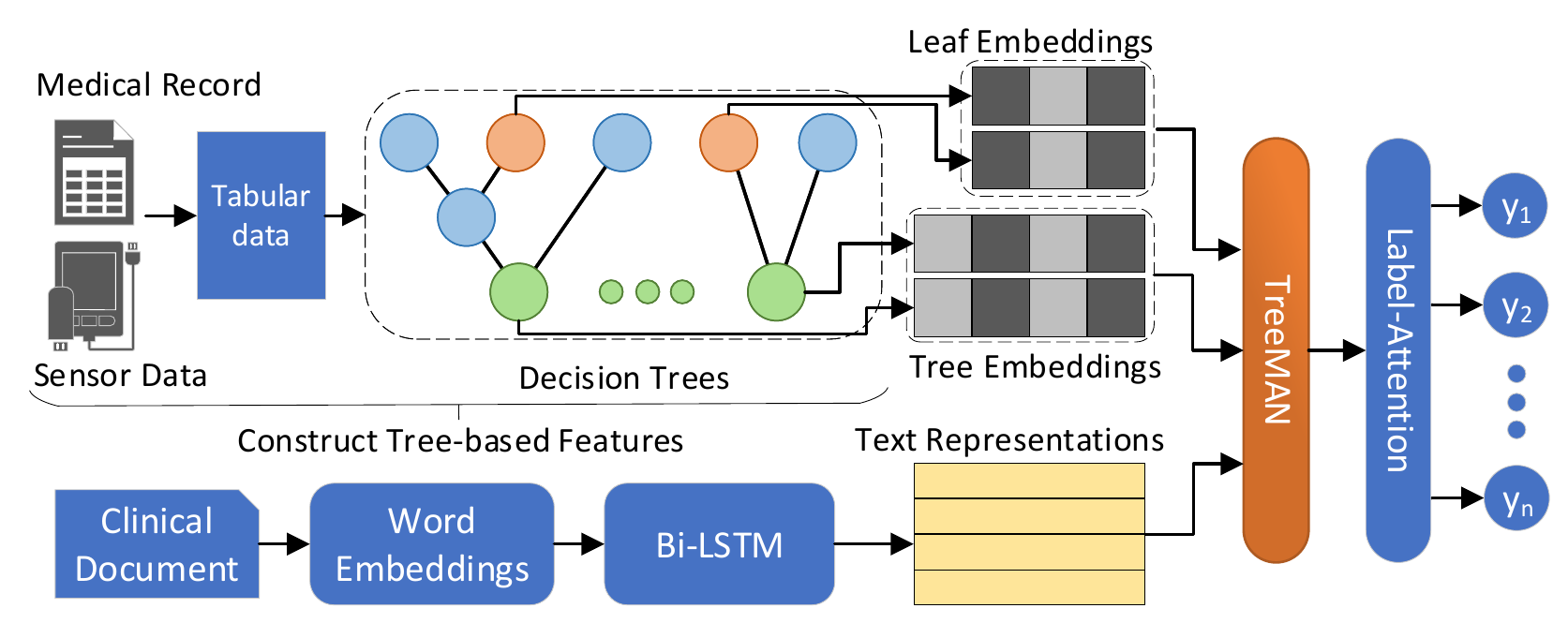}
  \caption{An overview of our proposed multimodal ICD coding framework. }\label{fig:overview}
\end{figure*}

\subsection{Tree-based Method}

Decision trees are a supervised learning algorithm broadly applied in regression and classification tasks \citep{10.1023/A:1022643204877}. They are trained on labeled data while 
requiring little data preparation and domain knowledge while the preprocessed features are able to be fused with text representations easily.
Multiple skills have been implemented to ensemble relatively simple decision trees to get better performance \citep{DBLP:journals/pami/BanfieldHBK07,4796917}. Among all of these ideas,  
Gradient boosting decision tree (GBDT) is an important instance which introduces iterative functional gradient descent algorithms to boosting models firstly   \citep{10.1214/aos/1013203451}. Significant improvement made by XGBoost \citep{10.1145/2939672.2939785} and LightGBM \citep{NIPS2017_6449f44a} which use different gradient information to improve accuracy and training efficiency respectively. Many attempts \citep{10.1145/2351356.2351358, 10.1145/3041021.3054192} have been made based on decision-tree boosting algorithm since people found it could generate interpretable and effective cross-feature and is easy to fix with other models. A combination of GBDT with linear model like Logistic Regression(LR) effectively helps the models to make explainable predictions by selecting top cross features \citep{DBLP:conf/kdd/HePJXLXSAHBC14}. \citet{DBLP:conf/www/Wang0FNC18} argued that the tree-enhanced embedding method would benefit from the explainability of tree-based models, thus improving generalization ability compared with other pure embedding ways. 
Incorporating decision tree learning with matrix factorization would help to extract the latent factors and get Fine-Grained embedding with rich semantic information  \citep{10.1145/3394486.3403339}, which could contribute to solve cold-start problems even further \citep{10.1145/3331184.3331244,10.1145/2009916.2009961}.

\begin{figure*}[t]
  \centering
  \includegraphics[width=0.85\textwidth]{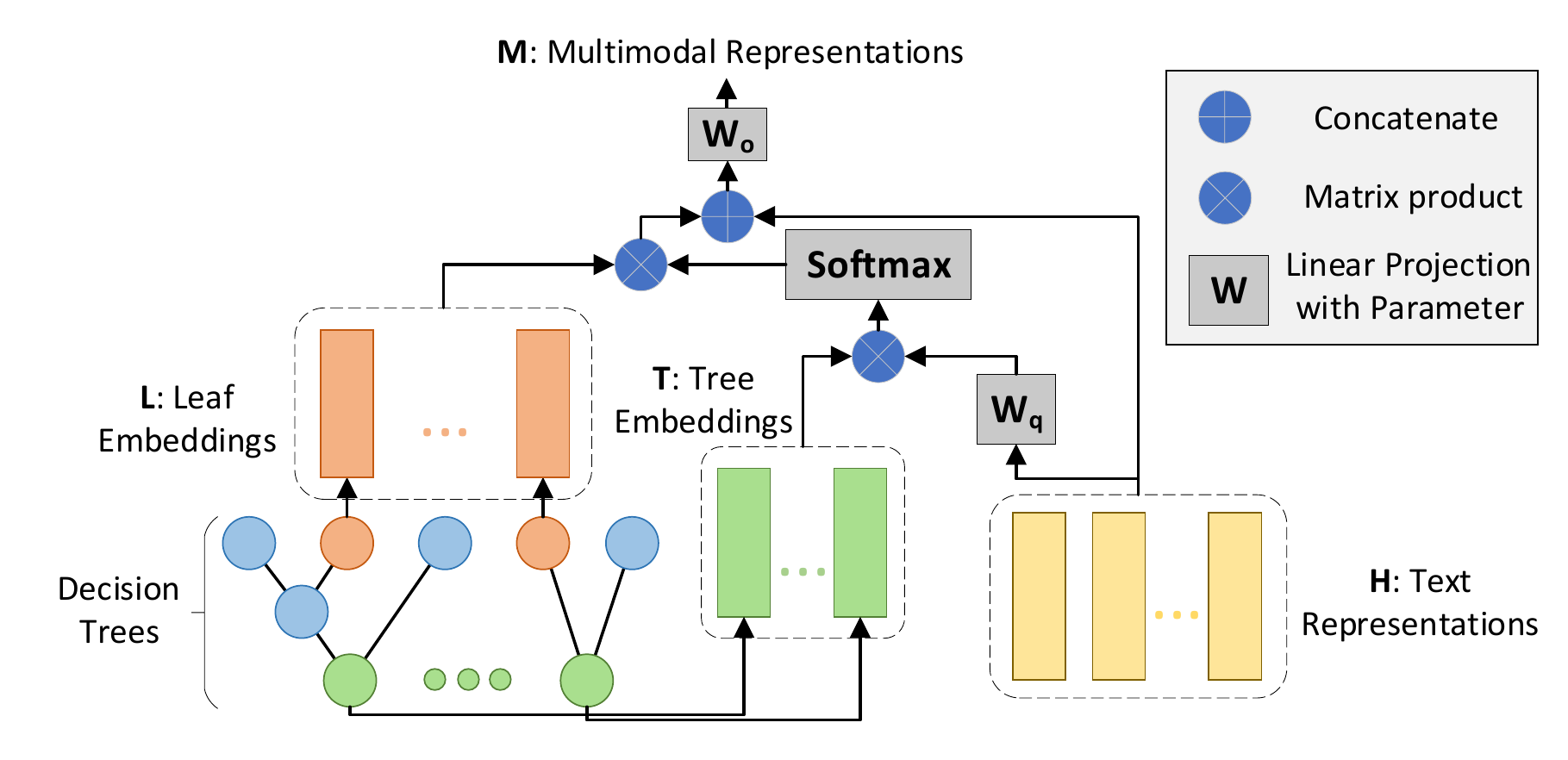}
  \caption{An illustration of our Tree-enhanced Multimodal Attention Network (TreeMAN). Green nodes and orange nodes on the decision trees respectively represent root nodes and activated leaf nodes. }\label{fig:att}
\end{figure*}

\section{Method}\label{sec:method}

In this section, we first give an overview of our framework (Section \ref{sec:over}),
and then detail the key module in our framework: the tree-enhanced multimodal attention network TreeMAN (Section \ref{sec:treeman}).
Finally, we introduce the processing of structured medical data and decision tree learning (Section \ref{sec:cross}).

\subsection{Overview}\label{sec:over} 

Figure \ref{fig:overview} shows the overview of our method. 
Upon discharge, there are two types of data available for our model: clinical notes written by doctors and structured medical data, including physiological data collected by sensors and medical records, such as lab measurements and prescriptions.
Given a clinical note and the associated structured data, two modules in the model process them separately to obtain text representations and tree-based features.
Considering in the poor performance of Bert-
like models on ICD coding\citep{DBLP:conf/emnlp/ChalkidisFKMAA20,DBLP:journals/corr/abs-2006-03685}, we train the text model from
scratch instead of fine-tuning a pretrained language
model. 
Structured medical data is first processed as tabular data and then fed into a trained decision tree to obtain tree-based features that we project into embedding vectors: the tree embeddings $\textbf{T}$ and the leaf embeddings $\textbf{L}$ (detailed in Section \ref{sec:cross}).
The other module is the text encoder designed to capture the semantic information in the document and provide textual representations.

\paragraph{Text encoder}
Given an input document with $N$ words $\{w_i\}^N_{i=1}$, the encoder first maps each word $w_i$ to a $d_e$-dimensional pre-trained word embedding $\textbf{e}_i$, then concatenates embeddings into the matrix $\textbf{E}=[\textbf{e}_1, \textbf{e}_2, ..., \textbf{e}_N]$. %
To capture contextual information, the word embedding matrix $E$ is fed into a bidirectional LSTM layer to compute the text representations $\textbf{H}$, which is a concatenation of the forward output and the backward output:
\begin{equation}
    \begin{gathered}
        \overrightarrow{\textbf{h}_i} = \overrightarrow{LSTM}(\textbf{e}_{1:i}), \quad
        \overleftarrow{\textbf{h}_i} = \overleftarrow{LSTM}(\textbf{e}_{i:N}), \\
        \textbf{H} = [\overrightarrow{\textbf{h}_1} \oplus \overleftarrow{\textbf{h}_1} ,\overrightarrow{\textbf{h}_2} \oplus \overleftarrow{\textbf{h}_2}
        ,...,\overrightarrow{\textbf{h}_n} \oplus \overleftarrow{\textbf{h}_n}]. 
    \end{gathered}
\end{equation}

Then, text representations $\textbf{H}$ together with the tree embeddings $\textbf{T}$ and the leaf embeddings $\textbf{L}$ generated from tree-based features are fed to TreeMAN to obtain the multimodal representation $\textbf{M}$ (detailed in Section \ref{sec:treeman}):
\begin{equation}
    \textbf{M} = \text{TreeMAN}(\textbf{H}, \textbf{L}, \textbf{T}).
\end{equation}

In the output layer, following \citet{DBLP:conf/naacl/MullenbachWDSE18}, we apply the per-label attention network to compute representations for each label. 

\paragraph{Label attention}
The label attention network takes multimodal representations $\textbf{M} \in \mathbb{R}^{d_m \times N}$ as input and compute the per-label representations $\textbf{V} \in \mathbb{R}^{d_m \times |\mathcal{L}|}$ with a matrix parameter $U \in \mathbb{R}^{d_m \times |\mathcal{L}|}$, where $|\mathcal{L}|$ represents the number of labels:
\begin{equation}
    \begin{split}
        \textbf{A} = \text{softmax}(\textbf{M}\textbf{U}), \\
        \textbf{V} = \textbf{A}^T\textbf{M}.
    \end{split}
\end{equation}

Finally, to compute the probability $\hat{y}_i$ of the $i^{th}$ label, the label representation $\textbf{v}_i$ of $\textbf{V}$ is fed into a corresponding linear layer followed by a sigmoid transformation.
For training, the model is optimized to minimize the binary cross-entropy loss between the prediction $\hat{y}$ and the target $y$:
\begin{equation}
    Loss=\sum_{i=1}^{|\mathcal{L}|}-y_ilog(\hat{y}_i)-(1-y_i)log(1-\hat{y}_i).
\end{equation}

\subsection{TreeMAN}\label{sec:treeman}

TreeMAN, a tree-enhanced multimodal attention network, is designed to fuse tree-based features and text representations and provides enhanced multimodal representations for multi-label classification.
We argue that the critical information in text representations is respective and fragmented because decisive information for different labels in the document is likely contained in different short segments \citep{DBLP:conf/naacl/MullenbachWDSE18}.
Therefore, we use the attention mechanism to learn the relevant features for each text vector and then fuse tree-based features and text information into a unified multimodal representation.

An illustration of TreeMAN is shown in Figure \ref{fig:att}.
Specifically, for each text vector $\textbf{h}_i \in \mathbb{R}^{d_h}$ in $\textbf{H}$, we first project it to a query vector $\textbf{q}_i$ by a learnable parameter, $\textbf{W}_q \in \mathbb{R}^{d_t \times d_h}$:

\begin{equation}
    \textbf{q}_i = \textbf{W}_q \textbf{h}_i.
\end{equation}

The vector $\textbf{q}_i \in \mathbb{R}^{d_t}$ is used to generate the attention weight $\bm{\alpha}_i$ by computing with the tree embeddings $\textbf{T} \in \mathbb{R}^{d_t \times |\mathcal{T}|}$, where $|\mathcal{T}|$ represents the number of decision trees:
\begin{equation}
    \bm{\alpha}_i = \text{softmax}(\textbf{T}^T\textbf{q}_i).
\end{equation}
The attention vector $\bm{\alpha}_i$ is then multiplied with the leaf embeddings $\textbf{L} \in \mathbb{R}^{d_l \times |\mathcal{T}|}$ to produce the special representation $\textbf{s}_i$ for the text vector:
\begin{equation}\label{equ:si}
    \textbf{s}_i = \textbf{L}\bm{\alpha}_i.
\end{equation}

To fuse the text information and tree-based features, we concatenate the special representation with text vector and then apply a linear projection:
\begin{equation}
    \textbf{m}_i = \textbf{W}_o[\textbf{h}_i||\textbf{s}_i],
\end{equation}
where $\textbf{W}_o \in \mathcal{R}^{d_m \times (d_l+d_s)}$ is a learnable parameter.
All the multimodal vectors are concatenated to formulate the output matrix $\textbf{M} = [\textbf{m}_1, \textbf{m}_2, ..., \textbf{m}_N] \in \mathbb{R}^{d_m \times N}$.

\subsection{Construction of Tree-based Features}\label{sec:cross}

In this section, we introduce how we construct the tree-based features from structured medical data.
Based on the characteristics of the data, we divide the structured medical data into three types: 
1) \textit{derived time series data} such as perioperative vital sign signals;
2) \textit{multivalued vertical data} denotes data with multiple records for one admission, such as lab measurements and prescriptions;
3) \textit{single horizontal data} indicates data with only a single record for an admission, such as admission type and patient age.
We process different types of data into tabular data in different ways: 
1) for \textit{derived time series data}, we compute mean, maximum, and minimum values for each class of data;
2) for \textit{multivalued vertical data}, we convert it into binary vector to indicate whether a test is abnormal or a medication is prescribed;
3) for \textit{single horizontal data}, we directly put it into the table as it is.

Then, we use the processed tabular data to construct decision trees by applying decision trees, which are trained with ICD codes as the target, using one-versus-all strategy for multi-label classification.

Formally, we get a set of decision trees, $\mathbb{Q} = \{Q_1, ..., Q_{|\mathcal{T}|}\}$, where each tree maps the tabular data $\textbf{x}$ to a leaf node, which can be represented by a one-hot vector.
The representation of tree-based features is a multi-hot vector $\textbf{q}$ which is a concatenation of one-hot vectors:
\begin{equation}
    \textbf{q} = [Q_1(x), ..., Q_{|\mathcal{T}|}(x)].
\end{equation}
Therefore, there are $|\mathcal{T}|$ elements of value 1 in $\textbf{q}$ indicates activated leaf nodes.

Inspired by the success of TEM \citep{DBLP:conf/www/Wang0FNC18}, we project $\textbf{q}$ into an embedding matrix $\textbf{L}$ as leaf embeddings.
For the attention computation in Section \ref{sec:treeman}, we also generate a tree embedding matrix $\textbf{T}$ based on the number of decision trees.

\section{Experiments}\label{sec:exp}

\begin{table}[t]
    \small
    \centering
    \begin{tabular}{p{2.4cm}p{1.9cm}p{1.9cm}}
    \hline
    \textbf{ } & \textbf{MIMIC-III 50}& \textbf{MIMIC-II 50}\\
    \hline
    Vocubulary Size  & 51,917 & 30,688\\
     \#  Samples & 11,371  & 3,726\\
    \hline
     \# *Drugs & 2350 & 52 \\
     \#  *Lab Items & 245 & 217 \\
     \#  *Organism & 183 & 135 \\
     \#  *Specimen & 74 & 63 \\
     \#  *Antibiotic & 30 & 30 \\
     \#  *Chart Items & 200 & - \\
    \hline
    Mean \# labels per document & 5.7 & 3.4\\
    Mean \# tokens per document & 1530 & 1014\\
    \hline
    \end{tabular}
    \caption{The statistics of the two MIMIC datasets and the structured medical data used therein, where "\#" indicates "the number of" and "*" denotes the number of classes is counted.}\label{Statistics}
\end{table}

\subsection{Datasets}
To make a fair and all-round comparison with former SOTA models, we evaluate our model on two widely used \textbf{M}edical \textbf{I}nformation \textbf{M}art for \textbf{I}ntensive \textbf{C}are (MIMIC) datasets: MIMIC-III  \citep{Johnson2016} and MIMIC-II \citep{1166854}. 
Because it's hard for our method to be implemented on ICD codes with less than 10 positive samples, we filter out records not relative to the top 50 most frequent ICD codes (denoted as MIMIC-III 50, MIMIC-II 50) to train and evaluate our method.

\paragraph{MIMIC-III 50.} 
Except for structured medical data, we use the same experimental setup including the same splits as previous works \citep{DBLP:conf/naacl/MullenbachWDSE18,DBLP:conf/acl/CaoCLZLC20,DBLP:conf/ijcai/VuNN20}.
For structured medical data, we use the following tables in MIMIC-III dataset
\footnote{A detailed introduction to MIMIC-III tables can be found at https://mimic.mit.edu/docs/iii/tables.} : 
1) \textit{Admissions} contains patients' admission information such as admission time; 
2) \textit{Patients} contains patients' basic information such as date of birth; 
3) \textit{Chartevents} contains charted data including patients’ routine vital signs; 
4) \textit{Labevents} contains laboratory measurements such as pH of blood;
5) \textit{Microbiologyevents} contains microbiology information such as organism test information;
6) \textit{Prescriptions} contains medications related to order entries including the Generic Sequence Number (GSN) of drugs.

\paragraph{MIMIC-II 50.} \label{sec:datasetII}
We subset the MIMIC-II full used in previous works \citep{DBLP:conf/naacl/MullenbachWDSE18} based on the 50 most frequent labels and use a set of 3,726 admission samples in which there are 2980 samples for training, 373 for validation and 373 for test.
For structured medical data, we use the following tables in MIMIC-II dataset
\footnote{A detailed introduction to MIMIC-II tables can be found at https://archive.physionet.org/mimic2/UserGuide/UserGui-de.pdf .}:
1) \textit{Admissions};
2) \textit{Patients};
3) \textit{Medevents} is similar to \textit{Prescriptions} in MIMIC-III;
4) \textit{Labevents};
5) \textit{Microbiologyevents}.

\paragraph{} 
Basic statistic information of all the datasets shows on Table \ref{Statistics}.

\begin{table*}
    \centering
    \resizebox{\textwidth}{!}{
    \begin{tabular}{|p{1.8cm}|c|c|c|c|c|c|c|c|c|c|}
    \hline
    \multirow{3}{*}{Model} & \multicolumn{5}{c|}{MIMIC-III 50}  & \multicolumn{5}{c|}{MIMIC-II 50} \\ \cline{2-11} 
    
     & \multicolumn{2}{c|}{AUC} & \multicolumn{2}{c|}{F1} & \multirow{2}{*}{P@5} & \multicolumn{2}{c|}{AUC} & \multicolumn{2}{c|}{F1} & \multirow{2}{*}{P@3} \\ \cline{2-5} \cline{7-10} 
                        & macro &micro & macro  & micro  &               & macro  & micro & macro & micro   &            \\
     \hline
            CAML        &0.875 &0.909  &0.532   &0.614  &0.609           &0.871  &0.902   &0.426   & 0.553  & 0.552     \\
            HyperCore   &0.895 &0.929  &0.609   &0.663  &0.632           & -     & -      & -      &-       &   -  \\
            LAAT        &0.925 &0.946  &0.666   &0.715  &0.675           &0.874  & 0.908  & 0.436  & 0.557  & 0.556      \\
            Joint LAAT  &0.925 &0.946  &0.661   &0.716  &0.671           &0.875  & 0.908  & 0.434  & 0.547  & 0.560    \\
            ISD         &0.935 &0.949  &0.679   &0.717  &\textbf{0.682}  &   -   &  -     & -       &    -    & -      \\
    \hline
      \multirow{2}{*}{\textbf{TreeMAN}}  &\textbf{0.937} &\textbf{0.953}  &\textbf{0.690} &\textbf{0.729} &\textbf{0.682}&\textbf{0.883}  &\textbf{0.916} & \textbf{0.479}  &\textbf{0.574} & \textbf{0.605}   \\
      & \small$ \pm 0.002$  &\small$ \pm 0.000$ & \small$ \pm 0.002$  &\small$ \pm 0.002$ & \small$ \pm 0.001$    &\small$ \pm 0.002$   &\small$ \pm 0.002$ &\small$ \pm 0.001$& \small$ \pm 0.001$   & \small$ \pm 0.004$                     \\
    \hline
    \end{tabular}
    }
    \caption{\label{result} Results on MIMIC-III 50 dataset, MIMIC-II 50 dataset and $mean \pm standard \,deviation$ of each indicator gained from replicated experiments with random initial states. Baseline scores are from the corresponding papers in Section \ref{sec::baseline}.} 
\end{table*}

\subsection{Implementation Details}

Following the preprocessing schema of previous works \citep{DBLP:conf/naacl/MullenbachWDSE18,DBLP:conf/aaai/Li020,DBLP:conf/cikm/XieXYZ19}, we lowercase all tokens and remove tokens that contain unrelated alphabetic characters like numbers and punctuations.
We implement the word2vec CBOW \citep{DBLP:journals/corr/MikolovSCCD13} method to pre-train word embeddings and  truncate  all discharge summary documents to the maximum length of 4,000 tokens. 
We employ XGBoost \footnote{https://xgboost.readthedocs.io.} to implement the decision trees in our approach.
There is only one decision tree built for each label where the learning rate and the maximum depth of the tree are set as 0.99 and 5, respectively
, while the rest of settings follow the default.
The sizes of the tree embedding $\textbf{T}$ and the leaf embedding $\textbf{L}$ are 128 and 30, respectively.
We set the size of multimodal representations to be the same as that of text representations.

For the baseline methods we reproduced on the MIMIC-II 50 dataset, we used the same implementations used by the authors on MIMIC-III 50.
To reduce randomness, we repeated all experiments 5 times with different random seeds and report the average performance.

\subsection{Metrics} 
To compare with previous and potential future work thoroughly, we measured our model mainly on indicators of macro-averaged and micro-averaged F1, macro-averaged, and micro-averaged AUC (area under the ROC curve) and Precision@k (P@k). 
Among these metrics, the “micro-averaged” method takes every single decision into consideration by pooling all text-code pairing and then calculating an effectiveness indicator on the pooled data.
And “macro-averaged based" metrics would provide statistics from the perspective of label instead of pair-relationship. Furthermore, we rank predictive probabilities to compute the precision of the top-k predicted labels, denoted as P@k. We set k to be five on MIMIC-III 50 dataset and three on MIMIC-II 50 dataset for the average discharge summary has 5.7 labels in  MIMIC-III 50 while 3.4 in MIMIC-II 50. We believe a full comparison of all the above metrics will provide insight into our work.

\subsection{Baselines} \label{sec::baseline}
We compare our model TreeMAN with the following baseline; all of them were SOTA when they were proposed initially.
\paragraph{CAML}  \textbf{C}onvolutional \textbf{A}ttention network for \textbf{M}ulti-
\textbf{L}abel classification (CAML) and \textbf{d}escription \textbf{R}egularized CAML was proposed by  \citet{DBLP:conf/naacl/MullenbachWDSE18}, which combined a single-layer CNN with attention layer to generate ICD coding for given text. 

\paragraph{LAAT\&Joint-LAAT}
\textbf{LA}bel \textbf{AT}tention  and  Joint \textbf{LA}bel \textbf{AT}tention
model was proposed by \citet{DBLP:conf/ijcai/VuNN20}. It encodes the input text with BiLSTM layer and implements self-attention mechanism to learn label-specific vectors representation. A hierarchical joint learning architecture is utilized to improve performance in the second model.

\paragraph{HyperCore}\textbf{Hyper}bolic$\,$and$\,$\textbf{Co}-graph Representation was proposed by \citet{DBLP:conf/acl/CaoCLZLC20}. It leveraged hierarchical structure of ICD code in hyperbolic space and used  graph convolutional network(GCN) to capture co-occurrence correlation of labels.

\paragraph{ISD} \textbf{I}nteractive \textbf{S}hared Representation Network with Self-\textbf{D}istillation Mechanism was proposed by \citet{DBLP:conf/acl/ZhouC000NCL20}, they implemented a self-distillation learning mechanism to alleviate the noisy text and only focus on noteworthy part of text.

\subsection{Results}
Table \ref{result} reports $mean \pm standard \ deviation$ of TreeMAN's results on two datasets, the performance of baselines on MIMIC-III 50 and the results of our implementation of baselines on MIMIC-II 50.
Compared with previous text methods, our multimodal approach achieves the best results on all metrics on both datasets.
It indicates that our model benefits from the rich information contained in structured medical data.
Furthermore, the small standard deviations demonstrate that the good results our model achieved are stable.

We also observe more significant improvements in the f1-marco and f1-micro metrics compared to other ranking-based metrics.
Since the binary output is produced by a fixed threshold $0.5$, a possible reason for the disparity is that the sigmoid function of our model in the final layer outputs more dispersed probabilities due to the decisive information provided by structured medical data.


\begin{table}
\centering
\begin{tabular}{p{1.7cm}|p{0.7cm}p{0.7cm}|p{0.7cm}p{0.7cm}|p{0.6cm}}
\hline
\multirow{2}{*}{Model} & \multicolumn{2}{c|}{AUC} & \multicolumn{2}{c|}{F1} & \multicolumn{1}{c}{\multirow{2}{*}{P@5}} \\ \cline{2-5}
 & macro & micro & macro & micro & \multicolumn{1}{c}{} \\ \hline
text & 92.6 & 94.5 & 67.4 & 71.4 & 66.6 \\
maxpooling & 93.1 & 94.9 & 68.4 & 72.3 & 67.5 \\
average & 93.4 & 95.1 & 68.9 & 72.7 & 67.6 \\ \hline \hline
TreeMAN & 93.7 & 95.3 & 69.0 & 72.9 & 68.2 \\ \hline
\end{tabular}
\caption{Results of ablation experiments on the MIMIC-III 50 dataset (in \%). \label{table:abla}}
\end{table}

\subsection{Ablation Experiment}
To testify the effectiveness of the different modules in TreeMAN, we perform a series of ablation experiments on the MIMIC-III 50 dataset, design following experiments, and report the results in Table \ref{table:abla}.

\paragraph{The Effect of Structured Medical Data} 
To study the effectiveness of the information captured from structured medical data,
we remove the tree-based features in TreeMAN and directly feed the unfused text representations to the multi-label classifier (\textit{text} in Table \ref{table:abla}).
The experimental results of all metrics decreased significantly compared to TreeMAN, demonstrating the importance of the tree-based features constructed based on structured medical data. 
It's also a comparison between the text representations and the multimodal representations, which proves that TreeMAN is capable of learning multimodal features.

\paragraph{The Effect of Attention Mechanism}
To examine the effectiveness of the attention mechanism in TreeMAN, we design two experiments by replacing the attention network with the max-pooling layer (\textit{maxpooling} in Table \ref{table:abla}) and the average layer (\textit{average} in Table \ref{table:abla}) on leaf embeddings. 
Formally, we change the Equation \ref{equ:si} as:
\begin{equation}
\left\{
    \begin{array}{lr}
        \text{average: } \textbf{s}_i = \frac{1}{|\mathcal{T}|} \sum_{\textbf{l} \in \textbf{L}}(\textbf{l}) ,\\
        \text{maxpooling: } \textbf{s}_i = max\_pool_{\textbf{l} \in \textbf{L}}(\textbf{l}) ,
    \end{array}
\right.
\end{equation}
where $\textbf{l}$ and $\mathcal{T}$ represent a vector in the leaf embeddings $\textbf{L}$ and the number of decision trees, respectively.
As shown, the experimental results of \textit{maxpooling} and \textit{average} are both better than \textit{text} and worse than TreeMAN.
Thus, the attention mechanism improves TreeMAN's ability to learn multimodal information and the information captured by tree-based features is robust to learn.

\subsection{Parameter Studies}

We have already analyzed the efficacy of our proposed model, and now we want to conduct a series of experiments to test the effect of two critical hyper-parameters in the TreeMAN module: the maximum depth of the decision tree and the size of leaf embeddings $\textbf{L}$. 
The former decides how tree-based features are constructed from structured medical data, 
and the latter is the representation format of the tree-based features. 
Various metrics of different settings would help us to demonstrate how TreeMAN extracts information from multimodal data:

\begin{figure*}[]
\centering
\subfigure[Performance of TreeMAN with different maximum Depth of Decision Tree\label{Depth}]{
\begin{minipage}[t]{0.47\linewidth}
\centering
\includegraphics[scale=0.47]{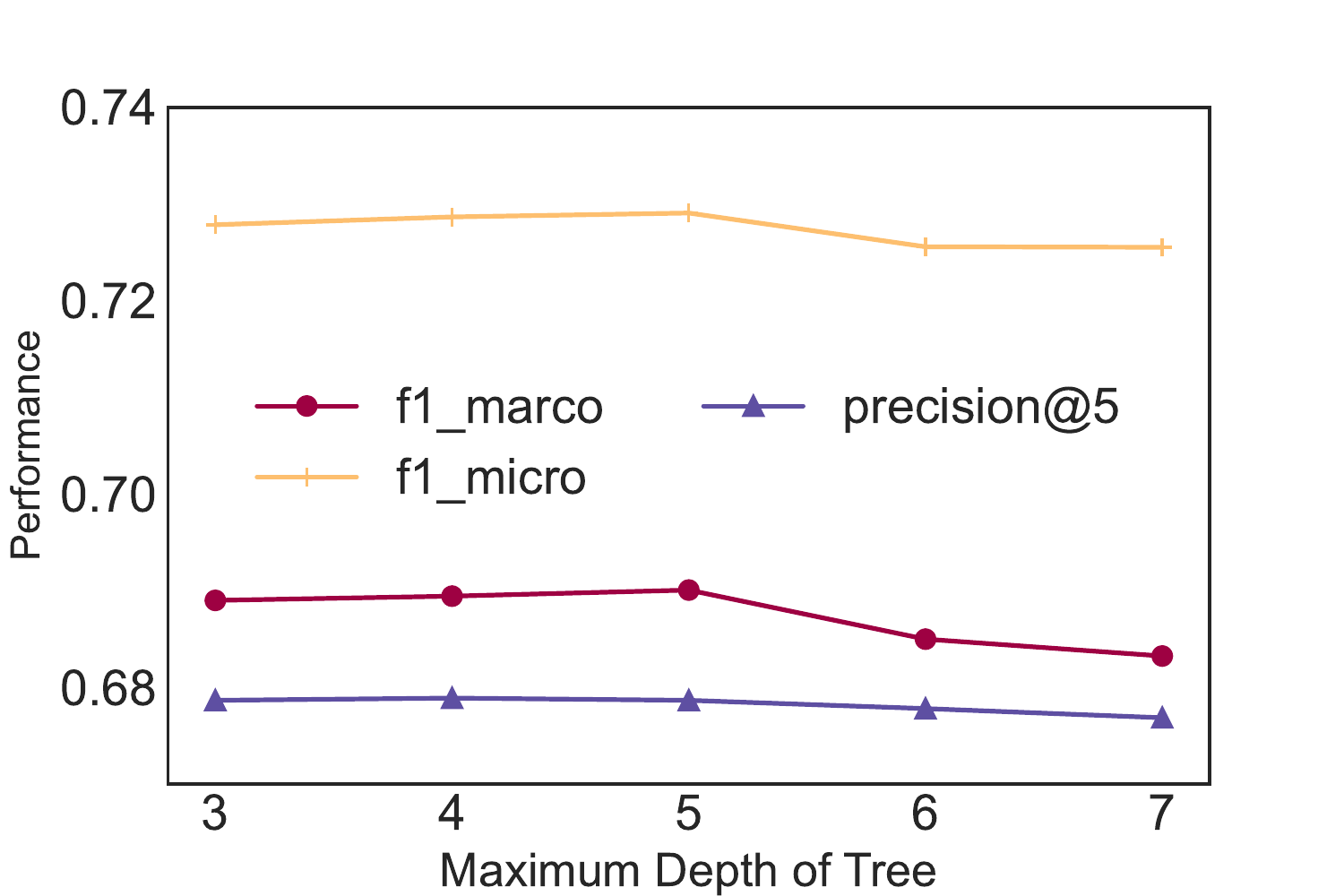}
\end{minipage}%
}%
\hspace{3mm}
\subfigure[Performance of TreeMAN with different Leaf Embedding Dimensions\label{label embedding}]{
\begin{minipage}[t]{0.47\linewidth}
\centering
\includegraphics[scale=0.47]{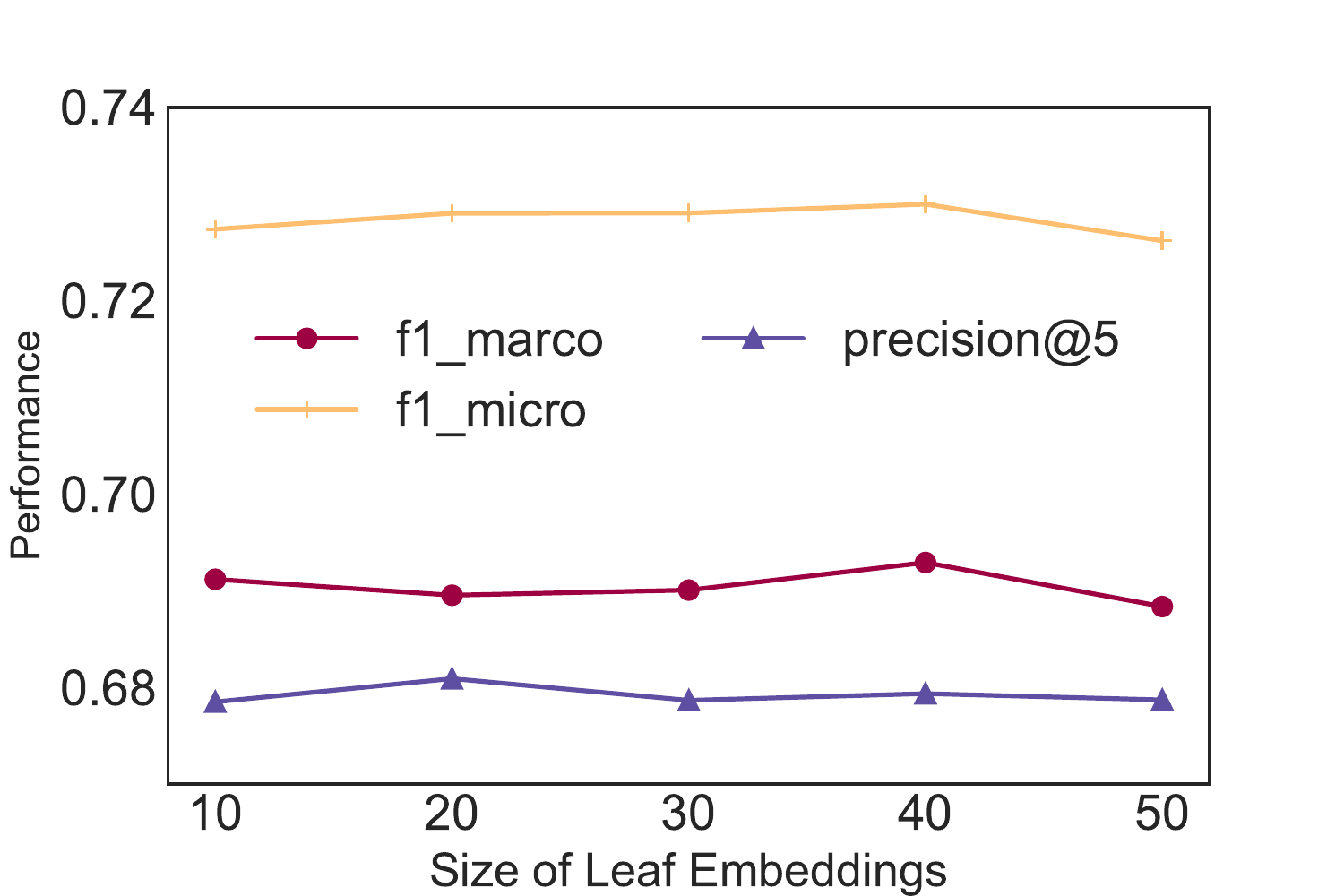}
\end{minipage}
}%
\centering
\caption{Results of different maximum tree depth and leaf embedding size on MIMIC-III 50\label{conclusionresult}.}
\end{figure*}

\paragraph{The Effect of Maximum Depth of the Decision Trees}

The  maximum depth of the decision tree would decide the number of feature extracted and the properties of the leaf embedding layer, for each of the leaf nodes represents a tree-based feature. For example, if we set the maximum depth to 3, we would get 401 leaves and 4752 leaves for a 7-layer tree. A shallow decision tree cannot extract enough features to represent the latent information of initial input. However, a too deep tree would risk over-fitting as well as colossal costs in the training process. 
Based on this assumption, we make a complete comparison of different pre-set depths of the decision tree. 
As the Figure \ref{conclusionresult} \subref{Depth} shows, a tree of depth 5 outperforms other decision trees, especially on the indicator of f1\verb|_|marco and f1\verb|_|micro because of the improvement in the aspect of recall ratio. 
Furthermore, we also notice that changes in this hyper-parameter don't seriously affect the performance of our module, proving the robustness of our method.

\paragraph{The Effect of Leaf Embeddings}

As we project multimodal information gained in the decision tree to leaf embeddings $\textbf{L}$,
we need the proper capacity of this layer to collect and store them. 
Thus we experiment with the leaf embedding size ranging from 10 to 50 to study the effect of the setup. 
Figure \ref{conclusionresult}\subref{label embedding}  shows that a vector with 30 dimensions is a proper choice because short vectors would abandon helpful information, while long ones would carry redundant information. Taking note of the limited size of datasets, relatively simple architecture could be a practical solution. These results also indicate that TreeMAN has learned an operative and steady pattern to learn from various types of multimodal information. 

\section{Conclusion}

In this paper, we proposed a tree-based multimodal method for the ICD coding task, which constructs tree-based features by decision trees learned from structured medical data and fuses the tree-based features and text representation by a novel tree-enhanced multimodal attention network (TreeMAN).
Experimental results on two MIMIC datasets show that our method outperforms state-of-the-art methods. 
Further ablation studies demonstrate that structured medical data and the attention mechanism in TreeMAN have improved the performance.

For future work, we plan to investigate the interpretability of our method since tree-based methods are naturally interpretable.
We are also interested in exploring a generalized and robust way to construct the tree-based features to capture more generalized medical information from structured medical data.

\section{Acknowledgement}
This research is supported by Chinese Scientific and Technical Innovation Project 2030 (No.2018AAA0102100), National Natural Science
Foundation of China (No. U1936206, 62077031, 62272250). We appreciate the reviewers for their constructive comments. We thank Shuyun Deng and Yuyang Shi
for their help in writing the paper.

\bibliography{custom}

\begin{thebibliography}{35}
\expandafter\ifx\csname natexlab\endcsname\relax\def\natexlab#1{#1}\fi

\bibitem[{Banfield et~al.(2007)Banfield, Hall, Bowyer, and
  Kegelmeyer}]{DBLP:journals/pami/BanfieldHBK07}
Robert~E. Banfield, Lawrence~O. Hall, Kevin~W. Bowyer, and W.~Philip
  Kegelmeyer. 2007.
\newblock \href {https://doi.org/10.1109/TPAMI.2007.250609} {A comparison of
  decision tree ensemble creation techniques}.
\newblock \emph{{IEEE} Trans. Pattern Anal. Mach. Intell.}, 29(1):173--180.

\bibitem[{Cai et~al.(2019)Cai, Wang, Li, and
  Liu}]{DBLP:journals/access/CaiWLL19}
Qiong Cai, Hao Wang, Zhenmin Li, and Xiao Liu. 2019.
\newblock \href {https://doi.org/10.1109/ACCESS.2019.2941419} {A survey on
  multimodal data-driven smart healthcare systems: Approaches and
  applications}.
\newblock \emph{{IEEE} Access}, 7:133583--133599.

\bibitem[{Cao et~al.(2020)Cao, Chen, Liu, Zhao, Liu, and
  Chong}]{DBLP:conf/acl/CaoCLZLC20}
Pengfei Cao, Yubo Chen, Kang Liu, Jun Zhao, Shengping Liu, and Weifeng Chong.
  2020.
\newblock \href {https://doi.org/10.18653/v1/2020.acl-main.282} {Hypercore:
  Hyperbolic and co-graph representation for automatic {ICD} coding}.
\newblock In \emph{Proceedings of the 58th Annual Meeting of the Association
  for Computational Linguistics, {ACL} 2020, Online, July 5-10, 2020}, pages
  3105--3114. Association for Computational Linguistics.

\bibitem[{Chalkidis et~al.(2020)Chalkidis, Fergadiotis, Kotitsas, Malakasiotis,
  Aletras, and Androutsopoulos}]{DBLP:conf/emnlp/ChalkidisFKMAA20}
Ilias Chalkidis, Manos Fergadiotis, Sotiris Kotitsas, Prodromos Malakasiotis,
  Nikolaos Aletras, and Ion Androutsopoulos. 2020.
\newblock An empirical study on large-scale multi-label text classification
  including few and zero-shot labels.
\newblock In \emph{{EMNLP} {(1)}}, pages 7503--7515. Association for
  Computational Linguistics.

\bibitem[{Chen and Guestrin(2016)}]{10.1145/2939672.2939785}
Tianqi Chen and Carlos Guestrin. 2016.
\newblock \href {https://doi.org/10.1145/2939672.2939785} {Xgboost: A scalable
  tree boosting system}.
\newblock In \emph{Proceedings of the 22nd ACM SIGKDD International Conference
  on Knowledge Discovery and Data Mining}, KDD '16, page 785–794, New York,
  NY, USA. Association for Computing Machinery.

\bibitem[{Ferr{\~a}o et~al.(2021)Ferr{\~a}o, Oliveira, Janela, Martins, and
  Gartner}]{ferrao2021can}
Jos{\'e}~Carlos Ferr{\~a}o, M{\'o}nica~Duarte Oliveira, Filipe Janela,
  Henrique~MG Martins, and Daniel Gartner. 2021.
\newblock Can structured ehr data support clinical coding? a data mining
  approach.
\newblock \emph{Health Systems}, 10(2):138--161.

\bibitem[{Friedman(2001)}]{10.1214/aos/1013203451}
Jerome~H. Friedman. 2001.
\newblock \href {https://doi.org/10.1214/aos/1013203451} {{Greedy function
  approximation: A gradient boosting machine.}}
\newblock \emph{The Annals of Statistics}, 29(5):1189 -- 1232.

\bibitem[{Gashler et~al.(2008)Gashler, Giraud-Carrier, and Martinez}]{4796917}
Mike Gashler, Christophe Giraud-Carrier, and Tony Martinez. 2008.
\newblock \href {https://doi.org/10.1109/ICMLA.2008.154} {Decision tree
  ensemble: Small heterogeneous is better than large homogeneous}.
\newblock In \emph{2008 Seventh International Conference on Machine Learning
  and Applications}, pages 900--905.

\bibitem[{He et~al.(2014)He, Pan, Jin, Xu, Liu, Xu, Shi, Atallah, Herbrich,
  Bowers, and Candela}]{DBLP:conf/kdd/HePJXLXSAHBC14}
Xinran He, Junfeng Pan, Ou~Jin, Tianbing Xu, Bo~Liu, Tao Xu, Yanxin Shi,
  Antoine Atallah, Ralf Herbrich, Stuart Bowers, and Joaquin~Qui{\~{n}}onero
  Candela. 2014.
\newblock \href {https://doi.org/10.1145/2648584.2648589} {Practical lessons
  from predicting clicks on ads at facebook}.
\newblock In \emph{Proceedings of the Eighth International Workshop on Data
  Mining for Online Advertising, {ADKDD} 2014, August 24, 2014, New York City,
  New York, {USA}}, pages 5:1--5:9. {ACM}.

\bibitem[{Johnson et~al.(2016)Johnson, Pollard, Shen, Lehman, Feng, Ghassemi,
  Moody, Szolovits, Anthony~Celi, and Mark}]{Johnson2016}
Alistair~E.W. Johnson, Tom~J. Pollard, Lu~Shen, Li-wei~H. Lehman, Mengling
  Feng, Mohammad Ghassemi, Benjamin Moody, Peter Szolovits, Leo Anthony~Celi,
  and Roger~G. Mark. 2016.
\newblock \href {https://doi.org/10.1038/sdata.2016.35} {Mimic-iii, a freely
  accessible critical care database}.
\newblock \emph{Scientific Data}, 3(1):160035.

\bibitem[{Ke et~al.(2017)Ke, Meng, Finley, Wang, Chen, Ma, Ye, and
  Liu}]{NIPS2017_6449f44a}
Guolin Ke, Qi~Meng, Thomas Finley, Taifeng Wang, Wei Chen, Weidong Ma, Qiwei
  Ye, and Tie-Yan Liu. 2017.
\newblock \href
  {https://proceedings.neurips.cc/paper/2017/file/6449f44a102fde848669bdd9eb6b76fa-Paper.pdf}
  {Lightgbm: A highly efficient gradient boosting decision tree}.
\newblock In \emph{Advances in Neural Information Processing Systems},
  volume~30. Curran Associates, Inc.

\bibitem[{Kim et~al.(2020)Kim, Tsai, Singh, Choi, Ibok, Li, and
  Cha}]{10.1145/3394486.3403339}
Sundong Kim, Yu-Che Tsai, Karandeep Singh, Yeonsoo Choi, Etim Ibok, Cheng-Te
  Li, and Meeyoung Cha. 2020.
\newblock \href {https://doi.org/10.1145/3394486.3403339} {Date: Dual attentive
  tree-aware embedding for customs fraud detection}.
\newblock KDD '20, page 2880–2890, New York, NY, USA. Association for
  Computing Machinery.

\bibitem[{Larkey and Croft(1996)}]{DBLP:conf/sigir/LarkeyC96}
Leah~S. Larkey and W.~Bruce Croft. 1996.
\newblock \href {https://doi.org/10.1145/243199.243276} {Combining classifiers
  in text categorization}.
\newblock In \emph{Proceedings of the 19th Annual International {ACM} {SIGIR}
  Conference on Research and Development in Information Retrieval, SIGIR'96,
  August 18-22, 1996, Zurich, Switzerland (Special Issue of the {SIGIR}
  Forum)}, pages 289--297. {ACM}.

\bibitem[{Li and Yu(2020)}]{DBLP:conf/aaai/Li020}
Fei Li and Hong Yu. 2020.
\newblock \href {https://ojs.aaai.org/index.php/AAAI/article/view/6331} {{ICD}
  coding from clinical text using multi-filter residual convolutional neural
  network}.
\newblock In \emph{The Thirty-Fourth {AAAI} Conference on Artificial
  Intelligence, {AAAI} 2020, The Thirty-Second Innovative Applications of
  Artificial Intelligence Conference, {IAAI} 2020, The Tenth {AAAI} Symposium
  on Educational Advances in Artificial Intelligence, {EAAI} 2020, New York,
  NY, USA, February 7-12, 2020}, pages 8180--8187. {AAAI} Press.

\bibitem[{Ling et~al.(2017)Ling, Deng, Gu, Zhou, Li, and
  Sun}]{10.1145/3041021.3054192}
Xiaoliang Ling, Weiwei Deng, Chen Gu, Hucheng Zhou, Cui Li, and Feng Sun. 2017.
\newblock \href {https://doi.org/10.1145/3041021.3054192} {Model ensemble for
  click prediction in bing search ads}.
\newblock In \emph{Proceedings of the 26th International Conference on World
  Wide Web Companion}, WWW '17 Companion, page 689–698, Republic and Canton
  of Geneva, CHE. International World Wide Web Conferences Steering Committee.

\bibitem[{Mikolov et~al.(2013)Mikolov, Sutskever, Chen, Corrado, and
  Dean}]{DBLP:journals/corr/MikolovSCCD13}
Tom{\'{a}}s Mikolov, Ilya Sutskever, Kai Chen, Greg Corrado, and Jeffrey Dean.
  2013.
\newblock \href {http://arxiv.org/abs/1310.4546} {Distributed representations
  of words and phrases and their compositionality}.
\newblock \emph{CoRR}, abs/1310.4546.

\bibitem[{Mullenbach et~al.(2018)Mullenbach, Wiegreffe, Duke, Sun, and
  Eisenstein}]{DBLP:conf/naacl/MullenbachWDSE18}
James Mullenbach, Sarah Wiegreffe, Jon Duke, Jimeng Sun, and Jacob Eisenstein.
  2018.
\newblock \href {https://doi.org/10.18653/v1/n18-1100} {Explainable prediction
  of medical codes from clinical text}.
\newblock In \emph{Proceedings of the 2018 Conference of the North American
  Chapter of the Association for Computational Linguistics: Human Language
  Technologies, {NAACL-HLT} 2018, New Orleans, Louisiana, USA, June 1-6, 2018,
  Volume 1 (Long Papers)}, pages 1101--1111. Association for Computational
  Linguistics.

\bibitem[{Nguyen et~al.(2018)Nguyen, Truran, Kemp, Koopman, Conlan, O'Dwyer,
  Zhang, Karimi, Hassanzadeh, Lawley, and
  Green}]{DBLP:conf/amia/NguyenTKKCOZKHL18}
Anthony~N. Nguyen, Donna~L. Truran, Madonna Kemp, Bevan Koopman, David Conlan,
  John O'Dwyer, Ming Zhang, Sarvnaz Karimi, Hamed Hassanzadeh, Michael Lawley,
  and Damian~J. Green. 2018.
\newblock \href
  {https://knowledge.amia.org/67852-amia-1.4259402/t004-1.4263758/t004-1.4263759/2976939-1.4263877/2974160-1.4263874}
  {Computer-assisted diagnostic coding: Effectiveness of an nlp-based approach
  using {SNOMED} {CT} to {ICD-10} mappings}.
\newblock In \emph{{AMIA} 2018, American Medical Informatics Association Annual
  Symposium, San Francisco, CA, November 3-7, 2018}. {AMIA}.

\bibitem[{O'malley et~al.(2005)O'malley, Cook, Price, Wildes, Hurdle, and
  Ashton}]{o2005measuring}
Kimberly~J O'malley, Karon~F Cook, Matt~D Price, Kimberly~Raiford Wildes,
  John~F Hurdle, and Carol~M Ashton. 2005.
\newblock Measuring diagnoses: Icd code accuracy.
\newblock \emph{Health services research}, 40(5p2):1620--1639.

\bibitem[{Park et~al.(2000)Park, Kim, Lee, Lee, Lee, Lee, Jee, Suh, Koh, Ryu
  et~al.}]{park2000accuracy}
Jong-Ku Park, Ki-Soon Kim, Tae-Yong Lee, Kang-Sook Lee, Duk-Hee Lee, Sun-Hee
  Lee, Sun-Ha Jee, Il~Suh, Kwang-Wook Koh, So-Yeon Ryu, et~al. 2000.
\newblock The accuracy of icd codes for cerebrovascular diseases in medical
  insurance claims.
\newblock \emph{Journal of Preventive Medicine and Public Health},
  33(1):76--82.

\bibitem[{Perotte et~al.(2014)Perotte, Pivovarov, Natarajan, Weiskopf, Wood,
  and Elhadad}]{perotte2014diagnosis}
Adler Perotte, Rimma Pivovarov, Karthik Natarajan, Nicole Weiskopf, Frank Wood,
  and No{\'e}mie Elhadad. 2014.
\newblock Diagnosis code assignment: models and evaluation metrics.
\newblock \emph{Journal of the American Medical Informatics Association},
  21(2):231--237.

\bibitem[{Quinlan(1986)}]{10.1023/A:1022643204877}
J.~R. Quinlan. 1986.
\newblock \href {https://doi.org/10.1023/A:1022643204877} {Induction of
  decision trees}.
\newblock \emph{Mach. Learn.}, 1(1):81–106.

\bibitem[{Rajendran et~al.(2021)Rajendran, Zenonos, Spear, and
  Pope}]{DBLP:conf/pkdd/RajendranZSP21}
Pavithra Rajendran, Alexandros Zenonos, Joshua Spear, and Rebecca Pope. 2021.
\newblock \href {https://doi.org/10.1007/978-3-030-93733-1\_26} {Embed wisely:
  An ensemble approach to predict {ICD} coding}.
\newblock In \emph{Machine Learning and Principles and Practice of Knowledge
  Discovery in Databases - International Workshops of {ECML} {PKDD} 2021,
  Virtual Event, September 13-17, 2021, Proceedings, Part {II}}, volume 1525 of
  \emph{Communications in Computer and Information Science}, pages 371--389.
  Springer.

\bibitem[{Saeed et~al.(2002)Saeed, Lieu, Raber, and Mark}]{1166854}
M.~Saeed, C.~Lieu, G.~Raber, and R.G. Mark. 2002.
\newblock \href {https://doi.org/10.1109/CIC.2002.1166854} {Mimic ii: a massive
  temporal icu patient database to support research in intelligent patient
  monitoring}.
\newblock In \emph{Computers in Cardiology}, pages 641--644.

\bibitem[{Safavian and Landgrebe(1991)}]{DBLP:journals/tsmc/SafavianL91}
S.~Rasoul Safavian and David~A. Landgrebe. 1991.
\newblock \href {https://doi.org/10.1109/21.97458} {A survey of decision tree
  classifier methodology}.
\newblock \emph{{IEEE} Trans. Syst. Man Cybern.}, 21(3):660--674.

\bibitem[{Tao et~al.(2019)Tao, Jia, Wang, and Wang}]{10.1145/3331184.3331244}
Yiyi Tao, Yiling Jia, Nan Wang, and Hongning Wang. 2019.
\newblock \href {https://doi.org/10.1145/3331184.3331244} {The fact: Taming
  latent factor models for explainability with factorization trees}.
\newblock In \emph{Proceedings of the 42nd International ACM SIGIR Conference
  on Research and Development in Information Retrieval}, SIGIR'19, page
  295–304, New York, NY, USA. Association for Computing Machinery.

\bibitem[{Teng et~al.(2022)Teng, Liu, Li, Zhang, Li, and Zhao}]{teng2022review}
Fei Teng, Yiming Liu, Tianrui Li, Yi~Zhang, Shuangqing Li, and Yue Zhao. 2022.
\newblock A review on deep neural networks for icd coding.
\newblock \emph{IEEE Transactions on Knowledge and Data Engineering}.

\bibitem[{Trofimov et~al.(2012)Trofimov, Kornetova, and
  Topinskiy}]{10.1145/2351356.2351358}
Ilya Trofimov, Anna Kornetova, and Valery Topinskiy. 2012.
\newblock \href {https://doi.org/10.1145/2351356.2351358} {Using boosted trees
  for click-through rate prediction for sponsored search}.
\newblock In \emph{Proceedings of the Sixth International Workshop on Data
  Mining for Online Advertising and Internet Economy}, ADKDD '12, New York, NY,
  USA. Association for Computing Machinery.

\bibitem[{Vu et~al.(2020)Vu, Nguyen, and Nguyen}]{DBLP:conf/ijcai/VuNN20}
Thanh Vu, Dat~Quoc Nguyen, and Anthony Nguyen. 2020.
\newblock \href {https://doi.org/10.24963/ijcai.2020/461} {A label attention
  model for {ICD} coding from clinical text}.
\newblock In \emph{Proceedings of the Twenty-Ninth International Joint
  Conference on Artificial Intelligence, {IJCAI} 2020}, pages 3335--3341.
  ijcai.org.

\bibitem[{Wang et~al.(2018)Wang, He, Feng, Nie, and
  Chua}]{DBLP:conf/www/Wang0FNC18}
Xiang Wang, Xiangnan He, Fuli Feng, Liqiang Nie, and Tat{-}Seng Chua. 2018.
\newblock \href {https://doi.org/10.1145/3178876.3186066} {{TEM:} tree-enhanced
  embedding model for explainable recommendation}.
\newblock In \emph{Proceedings of the 2018 World Wide Web Conference on World
  Wide Web, {WWW} 2018, Lyon, France, April 23-27, 2018}, pages 1543--1552.
  {ACM}.

\bibitem[{Xie et~al.(2019)Xie, Xiong, Yu, and Zhu}]{DBLP:conf/cikm/XieXYZ19}
Xiancheng Xie, Yun Xiong, Philip~S. Yu, and Yangyong Zhu. 2019.
\newblock \href {https://doi.org/10.1145/3357384.3357897} {{EHR} coding with
  multi-scale feature attention and structured knowledge graph propagation}.
\newblock In \emph{Proceedings of the 28th {ACM} International Conference on
  Information and Knowledge Management, {CIKM} 2019, Beijing, China, November
  3-7, 2019}, pages 649--658. {ACM}.

\bibitem[{Xu et~al.(2019)Xu, Lam, Pang, Gao, Band, Mathur, Papay, Khanna,
  Cywinski, Maheshwari, Xie, and Xing}]{DBLP:conf/mlhc/XuLPGBMPKCMXX19}
Keyang Xu, Mike Lam, Jingzhi Pang, Xin Gao, Charlotte Band, Piyush Mathur,
  Frank Papay, Ashish~K. Khanna, Jacek~B. Cywinski, Kamal Maheshwari, Pengtao
  Xie, and Eric~P. Xing. 2019.
\newblock \href {http://proceedings.mlr.press/v106/xu19a.html} {Multimodal
  machine learning for automated {ICD} coding}.
\newblock In \emph{Proceedings of the Machine Learning for Healthcare
  Conference, {MLHC} 2019, 9-10 August 2019, Ann Arbor, Michigan, {USA}},
  volume 106 of \emph{Proceedings of Machine Learning Research}, pages
  197--215. {PMLR}.

\bibitem[{Zhang et~al.(2020)Zhang, Liu, and
  Razavian}]{DBLP:journals/corr/abs-2006-03685}
Zachariah Zhang, Jingshu Liu, and Narges Razavian. 2020.
\newblock {BERT-XML:} large scale automated {ICD} coding using {BERT}
  pretraining.
\newblock \emph{CoRR}, abs/2006.03685.

\bibitem[{Zhou et~al.(2011)Zhou, Yang, and Zha}]{10.1145/2009916.2009961}
Ke~Zhou, Shuang-Hong Yang, and Hongyuan Zha. 2011.
\newblock \href {https://doi.org/10.1145/2009916.2009961} {Functional matrix
  factorizations for cold-start recommendation}.
\newblock In \emph{Proceedings of the 34th International ACM SIGIR Conference
  on Research and Development in Information Retrieval}, SIGIR '11, page
  315–324, New York, NY, USA. Association for Computing Machinery.

\bibitem[{Zhou et~al.(2021)Zhou, Cao, Chen, Liu, Zhao, Niu, Chong, and
  Liu}]{DBLP:conf/acl/ZhouC000NCL20}
Tong Zhou, Pengfei Cao, Yubo Chen, Kang Liu, Jun Zhao, Kun Niu, Weifeng Chong,
  and Shengping Liu. 2021.
\newblock \href {https://doi.org/10.18653/v1/2021.acl-long.463} {Automatic
  {ICD} coding via interactive shared representation networks with
  self-distillation mechanism}.
\newblock In \emph{Proceedings of the 59th Annual Meeting of the Association
  for Computational Linguistics and the 11th International Joint Conference on
  Natural Language Processing, {ACL/IJCNLP} 2021, (Volume 1: Long Papers),
  Virtual Event, August 1-6, 2021}, pages 5948--5957. Association for
  Computational Linguistics.

\end{thebibliography}
\bibliographystyle{acl_natbib}

\appendix



\end{document}